\documentclass[letterpaper, 10 pt, conference]{ieeeconf}  %

\IEEEoverridecommandlockouts                              %

\usepackage{graphics} %
\usepackage{epsfig} %
\usepackage{mathptmx} %
\usepackage{times} %
\usepackage{amsmath} %
\usepackage{amssymb}  %
\usepackage{amsfonts}
\usepackage{pgfplots}
\usepackage[utf8]{inputenc}
\usepackage{subcaption}
\usepackage{tikz}
\usepackage{siunitx}
\usepackage{pifont}
\usepackage{booktabs}
\usepackage{mathtools}
\usepackage{hyperref}
\usepackage[absolute,overlay]{textpos}
\usetikzlibrary{shapes,arrows, calc, arrows.meta, intersections, positioning, patterns} %
\definecolor{BOARD_GREEN}{RGB}{2,94,19}
\definecolor{BOARD_GREEN_2}{RGB}{58,112,57}
\definecolor{TRAINING}{RGB}{0,0,150}
\definecolor{CLOSED_LOOP}{RGB}{0,120,0}
\definecolor{INITIAL}{RGB}{170,0,0}

\DeclareMathAlphabet{\mathcal}{OMS}{cmsy}{m}{n}

\title{\LARGE \bf
True\AE dapt: Learning Smooth Online Trajectory Adaptation with Bounded Jerk, Acceleration and Velocity in Joint Space
}

\author{Jonas C. Kiemel$^{1*}$, Robin Weitemeyer$^{1*}$, Pascal Meißner$^{1}$ and Torsten Kröger%
\thanks{$^{1}$Institute for Anthropomatics and Robotics – Intelligent Process Automation and Robotics (IAR-IPR),
	Karlsruhe Institute of Technology (KIT),
	{\tt\small jonas.kiemel@kit.edu}  \newline $^{*}$ These authors contributed equally.}%
}	
\makeatletter
\tikzset{
    database top segment style/.style={draw},
    database middle segment style/.style={draw},
    database bottom segment style/.style={draw},
    database/.style={
        path picture={
            \path [database bottom segment style]
                (-\db@r,-0.5*\db@sh) 
                -- ++(0,-1*\db@sh) 
                arc [start angle=180, end angle=360,
                    x radius=\db@r, y radius=\db@ar*\db@r]
                -- ++(0,1*\db@sh)
                arc [start angle=360, end angle=180,
                    x radius=\db@r, y radius=\db@ar*\db@r];
            \path [database middle segment style]
                (-\db@r,0.5*\db@sh) 
                -- ++(0,-1*\db@sh) 
                arc [start angle=180, end angle=360,
                    x radius=\db@r, y radius=\db@ar*\db@r]
                -- ++(0,1*\db@sh)
                arc [start angle=360, end angle=180,
                    x radius=\db@r, y radius=\db@ar*\db@r];
            \path [database top segment style]
                (-\db@r,1.5*\db@sh) 
                -- ++(0,-1*\db@sh) 
                arc [start angle=180, end angle=360,
                    x radius=\db@r, y radius=\db@ar*\db@r]
                -- ++(0,1*\db@sh)
                arc [start angle=360, end angle=180,
                    x radius=\db@r, y radius=\db@ar*\db@r];
            \path [database top segment style]
                (0, 1.5*\db@sh) circle [x radius=\db@r, y radius=\db@ar*\db@r];
        },
        minimum width=2*\db@r + \pgflinewidth,
        minimum height=3*\db@sh + 2*\db@ar*\db@r + \pgflinewidth,
    },
    database segment height/.store in=\db@sh,
    database radius/.store in=\db@r,
    database aspect ratio/.store in=\db@ar,
    database segment height=0.1cm,
    database radius=0.25cm,
    database aspect ratio=0.35,
    database top segment/.style={
        database top segment style/.append style={#1}},
    database middle segment/.style={
        database middle segment style/.append style={#1}},
    database bottom segment/.style={
        database bottom segment style/.append style={#1}}
}
\makeatother

\tikzset{cross/.style={cross out, draw=black, minimum size=2*(#1-\pgflinewidth), inner sep=0pt, outer sep=0pt},
cross/.default={1pt}}

\newcommand*\rot{\rotatebox{90}}

\begin{document}

\maketitle
\thispagestyle{empty}
\pagestyle{empty}
\maxdeadcycles=20000

\begin{textblock*}{14.9cm}(3.2cm,0.75cm) %
	{\footnotesize © 2020 IEEE.  Personal use of this material is permitted.  Permission from IEEE must be obtained for all other uses, in any current or future media, including reprinting/republishing this material for advertising or promotional purposes, creating new collective works, for resale or redistribution to servers or lists, or reuse of any copyrighted component of this work in other works.}
\end{textblock*}

\begin{abstract}

We present True\AE dapt, a model-free method to learn online adaptations of robot trajectories based on their effects on the environment.
Given sensory feedback and future waypoints of the original trajectory, a neural network is trained to predict joint accelerations at regular intervals.
The adapted trajectory is generated by linear interpolation of the predicted accelerations, leading to continuously differentiable joint velocities and positions. 
Bounded jerks, accelerations and velocities are guaranteed by calculating the range of valid accelerations at each decision step and clipping the network's output accordingly. A deviation penalty during the training process causes the adapted trajectory to follow the original one. Smooth movements are encouraged by penalizing high accelerations and jerks. 
We evaluate our approach by training a simulated KUKA iiwa robot to balance a ball on a plate while moving and demonstrate that the balancing policy can be directly transferred to a real robot. A video presentation is available at \normalfont{\url{https://youtu.be/aNB0_tRxsuk}}.

\end{abstract}

\section{INTRODUCTION}

Robots frequently interact with their environment while executing movements. Industrial applications include spray painting, welding, bonding or grinding. 
In service robotics, an illustrative use-case is a waiter robot trying to transport glasses on a tray without spilling water.  

If the behaviour of the environment is precisely known in advance, motion planning can be performed offline. 
However, imperfect environment models or unforeseen external disturbances may cause the initial motion plan to fail.
For instance, welding distortion might be hard to predict, elastic components might cause problems during bonding and the grinding behaviour might alter over time due to wear of the abrasives. %
Reacting to unpredictable disturbances typically implies online adaptation of the initially planned trajectory. 
Designing a model-based control system for smooth trajectory adaptation in task space is challenging, especially if the robot is required to work near to kinematic singularities or close to the velocity limits of its joints to meet time requirements. 

With True\AE dapt, we replace the need for a plant model by learning how to adapt trajectories from simulated experiences using model-free reinforcement learning.
Kinematic singularities do not cause problems as the algorithm works in joint space. Bounded and continuously differentiable joint velocities are guaranteed and smooth adaptations are favoured since jerky movements are punished during training.  
We demonstrate successful sim-to-real transfer for a dynamic balancing task, which is motivated by the aforementioned job of a waiter robot. 
Like in the industrial applications mentioned above, the environment is directly influenced by the movements of the robot. 
However, the balancing task does not alter the environment permanently, which facilitates quantitative evaluation of the real-world performance.  
In addition, state-of-the-art physic engines allow fast simulation, making the task attractive for research on sim-to-real transfer.  

\begin{figure}[t]
\centering\
\begin{tikzpicture}
    \node[anchor=south west,inner sep=0] (image) at (0,0) {\includegraphics[trim=600 490 510 50, clip, width=\linewidth]{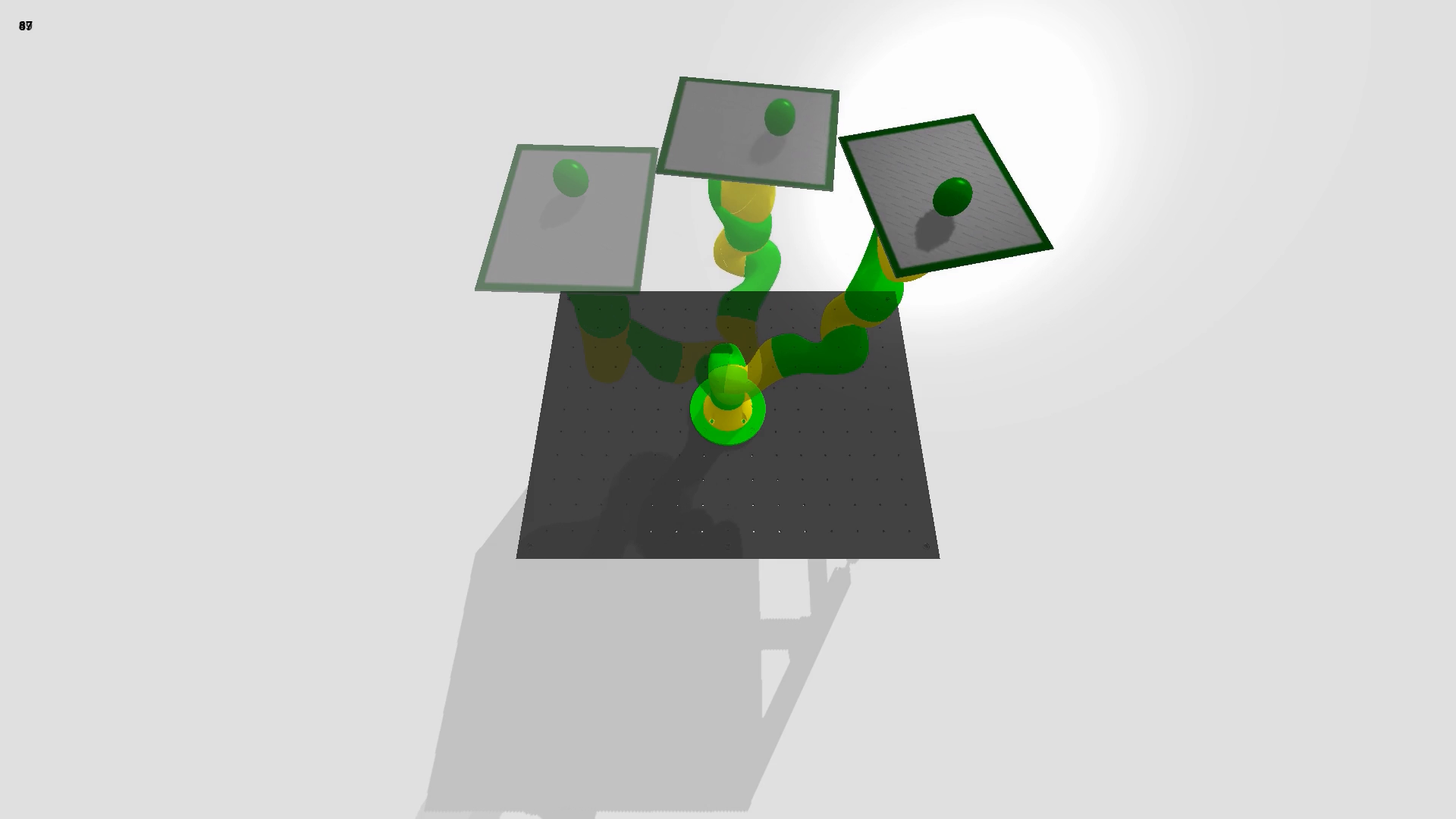}};
    \begin{scope}[x={(image.south east)},y={(image.north west)}]
        \draw[color=BOARD_GREEN_2, ultra thick, ->] (0.23, 0.83) arc (170:10:0.25 and 0.15);
        
    \end{scope}
    
\end{tikzpicture}
\caption{True\AE dapt applied to a balancing task: The robot has learned to keep a ball at the same spot on a plate while moving along a reference trajectory.}%
\label{fig:Closed_Loop_Dynamic}
\end{figure}

\tikzstyle{block} = [draw, rectangle, 
    minimum height=9em, minimum width=6em, align=center, text depth=20mm, text height=12mm]
\tikzstyle{sum} = [draw, fill=blue!20, circle, node distance=1cm]
\tikzstyle{input} = [coordinate]
\tikzstyle{output} = [coordinate]
\tikzstyle{pinstyle} = [pin edge={to-,thin,black}]

\tikzstyle{sysBlock} = [draw, rectangle, 
   minimum width=5.5em, minimum height=4.5em, rounded corners, align=center] 

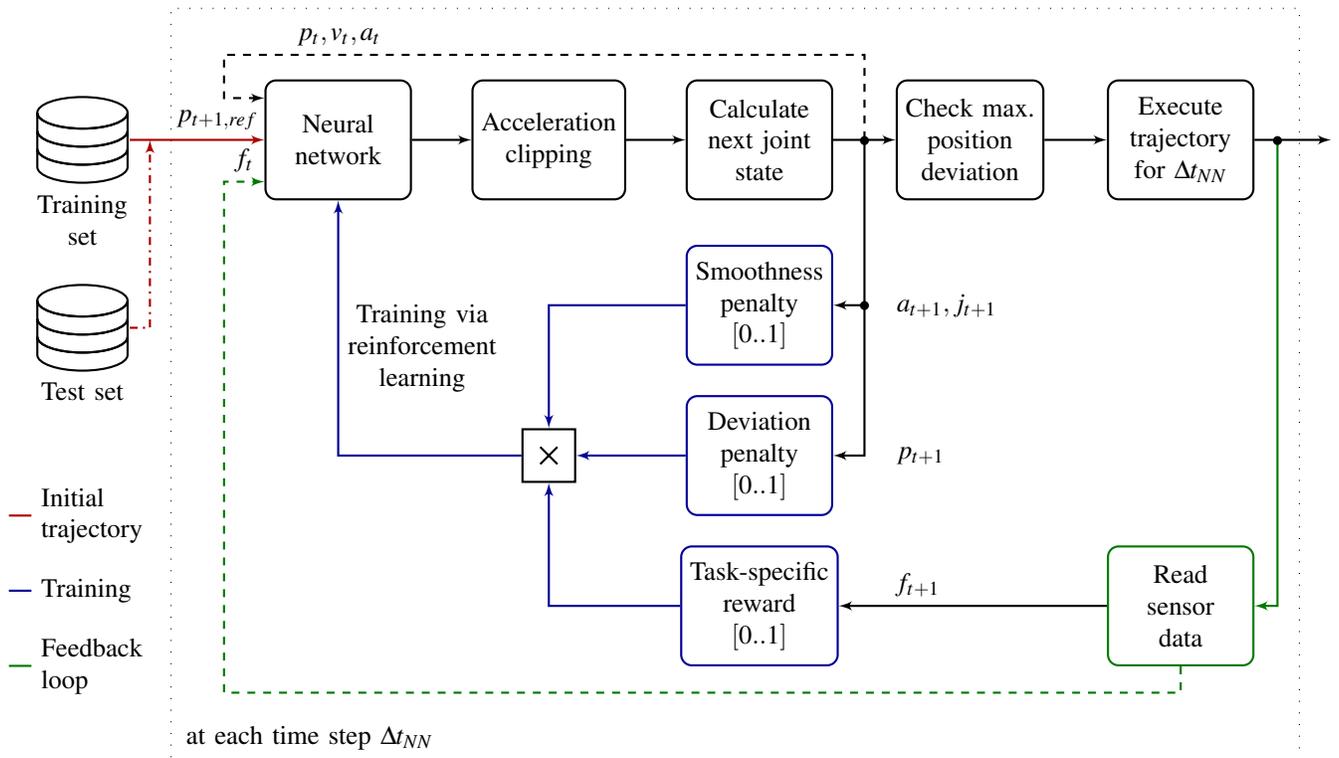
\begin{figure*}[h]
    \vspace{0.2cm}
	\centering
	\begin{tikzpicture}[auto, node distance=2.8cm,>=latex', thick]
    \node[database, label={[align=center]below:Training \\set},database radius=0.6cm,database segment height=0.24cm, database top segment={draw=black}](train){};
    \node[database,below of=train, node distance=2.5cm, label=below:Test set,database radius=0.6cm,database segment height=0.24cm, database top segment={draw=black}](test){};
    \node[draw=black, loosely dotted, thin, align=center, minimum height=10cm, minimum width=15cm] (forFrame) at ($ (train.east) +(8.05cm, -3.25cm) $) {};
    \node (forLoop) at ($(forFrame.south west) + (1.85 cm, 0.3cm)$) {at each time step $\Delta t_{NN}$}; %
    \node [sysBlock, right of=train, node distance=3.4cm] (network) {Neural \\ network};
    \node [sysBlock, right of=network] (clipping) {Acceleration \\ clipping};
    \node [sysBlock, right of=clipping] (setpoint) {Calculate \\ next joint \\ state};
    \node [sysBlock, right of=setpoint] (termination) {Check max. \\position \\deviation};
    \node [sysBlock, right of=termination] (execution) {Execute \\trajectory\\ for $\Delta t_{NN}$};
    \node [output, right of=execution, node distance=2cm] (output) {};
    
    \node [sysBlock, draw=TRAINING, below of=setpoint, node distance=2.2cm] (smooth) {Smoothness \\ penalty \\ $[0..1]$};
    \node [sysBlock, draw=TRAINING, below of=smooth, node distance=2cm] (dev) {Deviation \\ penalty \\ $[0..1]$};
    \node [sysBlock, draw=TRAINING, below of=dev, node distance=2cm] (reward) {Task-specific \\ reward \\ $[0..1]$};
    \node [sysBlock, draw=CLOSED_LOOP, right of=reward, node distance=5.6cm] (sensor) {Read \\ sensor \\ data};
    \node [draw, left of=dev, minimum height=0.7cm, minimum width=0.7cm](multiplication){}; 
    \def\mOff{0.25cm}
    \draw[black]($(multiplication.south east) + (-\mOff, \mOff)$) -- ($(multiplication.north west) + (\mOff, -\mOff)$) ($(multiplication.south west) + (\mOff, \mOff)$) -- ($(multiplication.north east) + (-\mOff, -\mOff)$);
    
    \def\inputOff{0.425cm}
    \draw [draw=INITIAL, ->] (train) -- node[pos=0.15, name=switch]{} 
    node[pos=0.65, above]{$p_{t+1, ref}$} node[pos=0.7, below=\inputOff, name=network_sensor_input]{}   node[pos=1, below=\inputOff, name=network_sensor_input2]{}  node[pos=0.7, above=\inputOff, name=network_joint_input]{}   node[pos=1, above=\inputOff, name=network_joint_input2]{} (network);
    \draw [draw=INITIAL, dashdotted, ->] (test) -| (switch);
    \draw [->] (network) -- (clipping);
    \draw [->] (clipping) -- (setpoint);
    \draw [->] (setpoint) -- node[pos=0.5] (setpoint_out) {}(termination);
    \fill ($(setpoint_out.center) - (0pt, 4pt)$) circle[radius=1.7pt];
    \draw [->] (termination) -- (execution);
    \draw [->] (execution) -- node[pos=0.3, name=execution_out] {}(output);
    
    \draw [draw=TRAINING, ->] (multiplication) -| node[pos=0.71, right, align=center]{Training via \\reinforcement \\ learning}  (network);
    \draw [draw=TRAINING, ->] (smooth) -| (multiplication);
    \draw [draw=TRAINING, ->] (dev) -- (multiplication);
    \draw [draw=TRAINING, ->] (reward) -| (multiplication);
    
    \draw [->] (setpoint_out) |- node[pos=0.5, right=0.3cm, align=center]{$a_{t+1}, j_{t+1}$} (smooth.0);
    \draw [->] ($(setpoint_out.center) - (0pt, 2.3cm)$)  |- node[pos=0.5, right=0.3cm, align=center]{$p_{t+1}$} (dev.0);
    \fill ($(setpoint_out.center) - (0pt, 2.3cm) - (0pt, 1pt)$) circle[radius=1.7pt];
    
    \draw [draw=CLOSED_LOOP, ->] (execution_out) |- (sensor.0);
    \fill ($(execution_out.center) - (0pt, 4pt)$) circle[radius=1.7pt];
    \draw [->] (sensor) -- node[pos=0.705, above]{$f_{t+1}$}(reward);
     \draw [draw=CLOSED_LOOP, dashed, ->] (sensor.south) -- ($(sensor.south) - (0pt, 0.35cm)$) -|  node[pos=0.94, right]{} (network_sensor_input.center) -- node[pos=0.5, above]{$f_{t}$} (network_sensor_input2.center);
     
     \draw [dashed, ->] ($(setpoint_out.center) - (0pt, 3.5pt)$) -- ($(setpoint_out.center) + (0pt, 1cm)$) -| node[pos=0.41, above, align=center]{$p_t, v_t, a_t$} (network_joint_input.center) -- (network_joint_input2.center);

     \draw [draw=INITIAL] ($(test.south west) + (-0.35cm, -1.9cm)$) --  node[pos=1, right, align=left]{Initial \\ trajectory}($(test.south west) + (-0.05cm, -1.9cm)$);
     \draw [draw=TRAINING] ($(test.south west) + (-0.35cm, -2.9cm)$) --  node[pos=1, right, align=left]{Training}($(test.south west) + (-0.05cm, -2.9cm)$);
     \draw [draw=CLOSED_LOOP] ($(test.south west) + (-0.35cm, -3.9cm)$) --  node[pos=1, right, align=left]{Feedback \\loop}($(test.south west) + (-0.05cm, -3.9cm)$);
    
\end{tikzpicture}
	\caption{System components to learn online adaptations with True\AE dapt. $\Delta t_{NN}$ is the time span between network predictions.}
	\label{fig:overview}
\end{figure*}

\section{Related Work}

\subsection{Trajectory Generation}
With sampling-based motion planners \cite{geraerts2004comparative}, finding a suitable robot trajectory is typically split in two distinct phases \cite{kunz2012time}.
Firstly, a collision-free geometric path is generated. Secondly, timestamps are added to the waypoints of the path, leading to a time-parameterized trajectory. 
The approach assumes that an appropriate path can be found without taking the timing of the movement into account. 
Although this assumption is not fulfilled for dynamic tasks like balancing, sampling-based motion planner can be used to generate reference trajectories for True\AE dapt. 
In \cite{Reflexxes}, a method for time-optimal online trajectory generation with bounded jerk and acceleration is presented. For offline scenarios, time-optimal trajectory parameterization can be performed considering both kinematic \cite{kunz2012time} and dynamic joint constraints \cite{toppra}. 

\subsection{Reinforcement Learning in Robotics}

In recent years, reinforcement learning (RL) has been applied to a variety of robotic applications like locomotion \cite{DBLP:journals/corr/abs-1812-11103, tan2018sim}, grasping \cite{berscheid_icra19, kalashnikov2018qt} or dexterous manipulation \cite{nagabandi2019deep, DBLP:journals_OpenAI_In_Hand}.
An RL-based method to smoothly track a jerky reference path with an industrial robot in the presence of unknown dynamical constraints is presented in \cite{TrajectoryOptimizationReferencePath}. The authors train a neural network to predict joint velocities and penalize the chosen action based on the distance to the reference path. In contrast, we predict joint accelerations to ensure continuously differentiable joint velocities and use a time-parameterized trajectory as reference. 
In \cite{levine2016end}, movements are learnt with a real robot by mapping a camera image directly to motor torques. 
However, when training in simulation, a very accurate dynamic model is required to generate meaningful torque commands for a real robot.

\subsection{Sim-to-real Transfer}
Generating sufficient training data for model-free RL-algorithms with real robots is costly and time-consuming. 
Conducting training in simulation is an appealing and widely used alternative. However, transfer from simulation to the real world typically leads to a drop in performance. %
One approach to bridge the so-called reality gap is randomization of the simulation to learn a robust policy. 
Domain randomization can be applied to simulated images \cite{tobin2017domain} as well as to physical parameters like friction or damping \cite{peng2018sim}. Making the simulation more realistic is another way to improve sim-to-real transfer. 
In \cite{graspGAN}, generative adversarial networks are trained to make synthetic renderings look like real images, whereas \cite{tan2018sim} incorporates an accurate actuator model and sensor latency to improve the simulation fidelity.

\section{System Overview}
The most important system components of True\AE dapt are shown in Fig. \ref{fig:overview}.
A neural network predicts joint accelerations based on sensory feedback, the current state of the joints and the following positions of a reference trajectory. The predicted accelerations are clipped to ensure that jerk, acceleration and velocity limits are not violated. In addition, the adapted trajectory is not executed if the adapted point deviates too much from the reference trajectory, thereby avoiding self-collision and violation of position limits. 
During training, a smoothness penalty penalizes jerky movements, while a deviation penalty ensures that the adapted trajectory follows the original one. A task-specific reward makes the system learn the intended task like balancing a ball. 
Details on each step will be explained in the following sections.

\section{Generation of Reference Trajectories}

\begin{figure}[h]
	\centering
	\begin{tikzpicture}[auto, node distance=3cm,>=latex']
    \node [block] (waypoint) {Task-specific \\ waypoint \\ generation
    };
    \node [block, right of=waypoint] (spline) {Spline \\ interpolation};
    \node [block, right of=spline] (parameterization) {Time-optimal \\ parameteri- \\zation};
    \node [block, below of=parameterization, 
            node distance=4cm, text height=16mm] (uniform) {Uniform \\ sampling \\ with \\ $\Delta t_{NN}$};
    \draw [draw,->] (waypoint) -- node[name=m] {}  (spline);
    \node[database,below of=m,node distance=2.95cm, label=below:Training set,database radius=0.85cm,database segment height=0.24cm, database top segment={draw=black,fill=gray!20}](train){};
    \node[database,below of=train, node distance=2cm, label=below:Test set,database radius=0.85cm,database segment height=0.24cm, database top segment={draw=black,fill=gray!20}](test){};

    \draw [->] (spline) -- (parameterization);
    \draw [->] (parameterization) -- (uniform);
    \draw [->] (uniform.180) --  (train.15);
    \draw [dashed, ->] (uniform.180) -- node[pos=0.58, above=0.28cm]{$\{p_{t, ref} \}_{t \in T}$} (test);

     \def\yWP{0.2cm}
    \draw[thick, draw=blue, dotted] ($ (waypoint.south west) + (0.2cm, 0.8cm+\yWP) $) rectangle ++(0.5cm,0.4cm);
     \draw[thick, draw=blue, dotted] ($ (waypoint.south west) + (0.9cm, 0.3cm+\yWP) $) rectangle ++(0.35cm,0.5cm);
    \draw[thick, draw=blue, dotted] ($ (waypoint.south west) + (1.45cm, 0.8cm+\yWP) $) rectangle ++(0.5cm,0.4cm);
    \draw ($ (waypoint.south west) + (1.075cm, 1.2cm+\yWP) $) circle [radius=0.1cm];
    \draw ($ (waypoint.south west) + (0.35cm, 1.0cm+\yWP) $) node[cross=2pt] {};
    \draw ($ (waypoint.south west) + (1.0cm, 0.5cm+\yWP) $) node[cross=2pt] {};
    \draw ($ (waypoint.south west) + (1.7cm, 0.95cm+\yWP) $) node[cross=2pt] {};
    
    \draw ($ (spline.south west) + (0.35cm, 1.0cm+\yWP) $) node[cross=2pt] (c1) {};
    \draw ($ (spline.south west) + (1.0cm, 0.5cm+\yWP) $) node[cross=2pt] (c2) {};
    \draw ($ (spline.south west) + (1.7cm, 0.95cm+\yWP) $) node[cross=2pt] (c3) {};
    \draw [blue] plot [smooth, tension=0.7] coordinates { (c1.center) (c2.center) (c3.center)};
    
    \begin{axis}[
          at={($(parameterization.south west) + (0.35cm, 0.4cm)$)},
          axis lines=middle,
          width=3.2cm,
          xmin=0,xmax=1,ymin=0,ymax=0.7,
          xtick distance=10,
          ytick distance=10,
          xlabel=$t$,
          ylabel={$p$},
          y label style={at={(axis description cs:0.15,1.1)},anchor=north},
          yticklabels={,,},
          xticklabels={,,},
          title={}
          ]
          grid style={thin,densely dotted,black!20}]
        \addplot[smooth,
                draw=blue,
                tension=1]
                coordinates{
                (0, 0.25) (0.2,0.35)
                (0.55,0.25) %
                (0.9, 0.35)
                };
    \end{axis}
    
    \begin{axis}[
          at={($(uniform.south west) + (0.35cm, 0.4cm)$)},
          axis lines=middle,
          width=3.2cm,
          xmin=0,xmax=1,ymin=0.0,ymax=0.7,
          xtick distance=10,
          ytick distance=10,
          xlabel=$t$,
          ylabel={$p$},
          y label style={at={(axis description cs:0.15,1.1)},anchor=north},
          yticklabels={,,},
          xticklabels={,,},
          title={}
          ]
          grid style={thin,densely dotted,black!20}]
        \addplot[smooth,
                name path=plot, 
                draw=black,
                tension=1]
                coordinates{
                (0, 0.25) (0.2,0.35)
                (0.55,0.25) %
                (0.9, 0.35)
                };
        \pgfplotsinvokeforeach{0.15,0.35,0.55, 0.75}{%
         \path[name path=a] (axis cs:#1, 0) -- (axis cs:#1,\pgfkeysvalueof{/pgfplots/ymax});
        
         \draw[blue, ->,
           name intersections={of=plot and a, total=\t,name=i}]
           \ifnum \t > 0
           (axis cs:#1,0) -- (i-1)
           \fi 
             ;
            }       
    \end{axis}
    
\end{tikzpicture}
	\caption{Generation of reference trajectories.}
	\label{fig:reference_generation}
\end{figure}
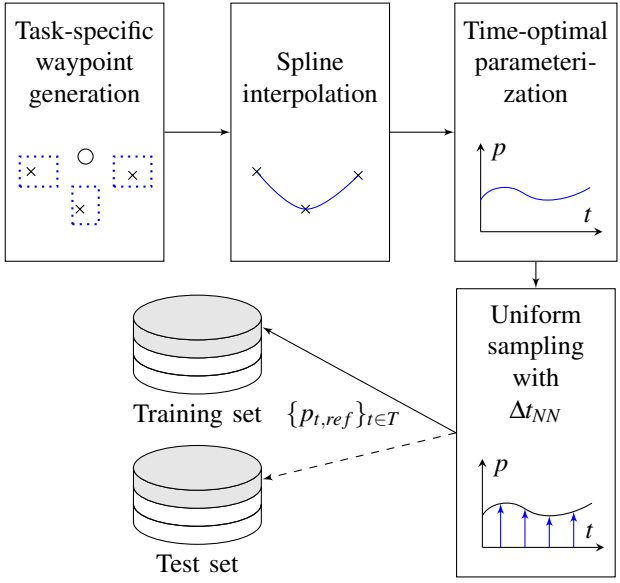

Suitable reference trajectories for True\AE dapt should follow the desired path of the movement, whereas dynamic interactions with the environment do not have to be considered.  
Our procedure to generate reference trajectories is illustrated in Fig. \ref{fig:reference_generation}.
As a first step, Cartesian waypoints are sampled randomly within predefined areas. 
Spline interpolation is used to produce a smooth Cartesian path. After converting the path to joint space via inverse kinematics, time-optimal trajectory parameterization is performed with a method described in \cite{kunz2012time}.
As a final step, the trajectory is uniformly sampled using the time span between network predictions $\Delta t_{NN}$, which we choose to be \SI{50}{\milli\second} for our experiments. 

Each trajectory is assigned either to the training set or to the test set.  

We note, that an offline method like \cite{TrueRMA} can be used to generate appropriate trajectories without the need to define task-specific sampling areas.

\section{Learning Online Trajectory Adaptations}

\subsection{Objectives}
We define the following objectives for our online trajectory adaptation approach:
\begin{itemize}
    \item The primary goal is to accomplish the specified task (e.g. balancing a ball).
    \item The adapted trajectory should stay close to the original one.
    \item Jerk, acceleration, velocity and position limits of the joints should not be violated.
    \item Self-collision should be avoided.
    \item The adapted trajectory should be smooth.
\end{itemize}

\subsection{Formalization}
The learning problem is formalized as a Markov Decision Process $(\mathcal{S}, \mathcal{A}, R_{\underline{a}})$, where $\mathcal{S}$ is the state space,  $\mathcal{A}$ is the action space and $R_{\underline{a}}$ is the immediate reward due to action $\underline{a}$. We use model-free RL for training a policy $\pi: \mathcal{S} \mapsto \mathcal{A}$ to maximize the expected sum of future rewards.
Each element of ${s \in \mathcal{S}}$  and $\underline{a} \in \mathcal{A}$ is normalized to be in the range of \mbox{[-1, 1]}. 
Decisions are made in real-time during motion with a cycle time of $\Delta t_{NN}$.
\subsubsection{State Definition}
The state ${s_t}$ consists of the current joint position $p_t$, velocity $v_t$ and acceleration $a_t$ as well as sensory feedback $f_t$ and $N$ future positions of the reference trajectory $p_{t+\{1..N\},ref}$.
Instead of using measured values for $p_t$, $v_t$ and $a_t$, we use the setpoints from the previous calculation step, thereby avoiding sensor noise and latency. 
The results of ablation studies to identify the influence of each part of the state can be found in TABLE \ref{table:evalution}. 
\subsubsection{Action Definition}

The action $\underline{a_t}$ determines $a_{t+1}$, the normalized angular acceleration for each robot joint at the beginning of the next time step.
Jerk, acceleration and velocity limits are respected by clipping the predicted acceleration $a_{t+1}$ accordingly. 
Linear interpolation between $a_{t}$ and $a_{t+1}$ is performed to produce continuous accelerations within the current time step.
Intermediate setpoints for a position controller are generated by integrating the accelerations twice.
We note that the movements of an untrained agent are not influenced by the selected reference trajectory.  
Instead, the network learns to stay close to the reference trajectory because of a deviation penalty.
With our approach, the execution times of the adapted trajectory and the reference trajectory are identical. 

\subsubsection{Reward Definition}
The reward per decision step ${R_{\underline{a}} \in [0, 1]}$ is calculated by multiplying a task-specific reward $R_T \in [0, 1]$ with a smoothness penalty $P_S \in [0, 1]$ and a deviation penalty $P_D \in [0, 1]$.   
\begin{align}
R_{\underline{a}} = R_T \cdot \left(1-P_S \right) \cdot \left(1-P_D \right) 
\end{align}
The smoothness penalty $P_S$ is composed of an acceleration penalty $P_A \in [0, 1]$ and a jerk penalty $P_J \in [0, 1]$.
\begin{align}
P_S = \frac{P_A + P_J}{2}
\end{align}
$P_A$ penalizes accelerations that are higher than a user-defined threshold $a_{th}$. In the following equation, $a_{abs} \in [0, 1]$ is the highest absolute value in $a_{t+1}$.
\begin{align}
P_A = \begin{dcases*}
0 & $a_{abs} \in [0, \; a_{th})$ \\
\left(1- \frac{1-a_{abs}}{1-a_{th}}\right)^2 & $a_{abs} \in [a_{th}, \; 1]$
\end{dcases*}
\end{align}
The following definition of $P_J$ is inspired by \cite{kalakrishnan2011stomp}. $N_J$ corresponds to the number of joint. $j_{abs, \; t+1, \; i}$ is the unnormalized absolute jerk of joint $i$, while $j_{abs, \; max, \; i}$ is the unnormalized jerk limit. $c$ is a user-defined weighting factor.
\begin{align}
j_{p} = \sum_{i=1}^{N_J} \left(j_{abs, \; t+1, \; i}\right)^2 \\
j_{sat} = \frac{1}{c} \cdot \sum_{i=1}^{N_J} \left(j_{abs, \; max, \; i}\right)^2 \\
P_J = \begin{dcases*}
\left(\frac{j_{p}}{j_{sat}} \right)^2 & $j_{p} \in [0, \; j_{sat}]$ \\
1 & $j_{p} > j_{sat}$
\end{dcases*}
\end{align}
The deviation penalty $P_D$ ensures that the adapted trajectory stays close to the reference. 
$\Delta p_{max}$ is the greatest absolute joint position deviation between $p_{t+1}$ and $p_{t+1, ref}$, while $\Delta p_{l}$ and $\Delta p_{h}$ are thresholds that lead to a punishment of 0 and 1, respectively.   
\begin{align}
P_D = \begin{dcases*}
0 & $\Delta p_{max} \in [0, \; \Delta p_{l})$ \\
\left(\frac{\Delta p_{max}-\Delta p_{l}}{\Delta p_{h} - \Delta p_{l}}\right)^2 & $\Delta p_{max} \in [\Delta p_{l}, \; \Delta p_{h}]$\\
1 & $\Delta p_{max}  > \Delta p_{h}$
\end{dcases*}
\end{align}

\begin{figure*}[t]
	\begin{subfigure}[c]{0.475\textwidth}
	    \vspace{0.05cm}
		\begin{tikzpicture}[scale=1.4]
		\draw [<-] (0,1.2) node (yaxis) [above] {$a$} -- (0,-0.8) node[below] {$t_0$};
		\draw [->] (-0.5, 0) -- (5,0) node (xaxis) [right] {$t$};
		\node at (2.25, -1.0) {\ldots};
		\draw [thick] (0,0.3) node[left] {$a_0$} -- (1.5,0.8) node[above right, yshift=-0.05cm] {$a_{max, v}$};
		\draw [thick](1.5,0.8) -- (4.5, -0.4) node[pos=0.23, below=0.04cm]{$j_{min}$}  node[pos=0.66,name=tazero]{} node[pos=0.66, below=0.05]{$t_{a_0}$} node[right] {$a_{n+1}$};
		\draw[thick] ($(tazero.center)+(0,2.5pt)$) -- ($(tazero.center)-(0,2.5pt)$);
		\fill[pattern=north east lines, pattern color=orange] (3.5, 0) --  (4.5, 0.0) -- (4.5, -0.4);
		\draw[dashed] (1.5, 1.2)  -- (1.5, -0.8) node[below] {$t_1$};
		\draw[dashed] (3, 1.2)  -- (3, -0.8) node[below] {$t_n$};
		\draw[dashed] (4.5, 1.2)  -- (4.5, -0.8) node[below] {$t_{n+1}$};
		\end{tikzpicture} 
		\subcaption{The maximum acceleration $a_{max, v}$ is followed by a deceleration with $j_{min}$. The maximum velocity $v_{max}$ is reached at $t_{a_0}$.}
		\label{fig:vel_limitation_a}
		
	\end{subfigure}
	\hspace{0.028\textwidth}
	\begin{subfigure}[c]{0.475\textwidth}
    	\vspace{0.05cm}
		\begin{tikzpicture}[scale=1.4]
		\draw [<-] (0,1.2) node (yaxis) [above] {$a$} -- (0,-0.8) node[below] {$t_0$};
		\draw [->] (-0.5, 0) -- (5,0) node (xaxis) [right] {$t$};
		\node at (2.25, -1.0) {\ldots};
		\draw [thick](0,0.3) node[left] {$a_0$} -- (1.5,0.75) node[above right] {$a_{max, v}^*$};
		\draw[densely dotted] (0,0.3)  -- (1.5,0.8);
		\draw[densely dotted, name path=init] (1.5,0.8) -- (4.5, -0.4);
		\draw [thick](1.5,0.75) -- (3, 0.15) node[pos=0.45, below=0.03cm]{$j_{min}$};
		\draw [name path=adapt, thick](3, 0.15) -- (4.5, 0) node[pos=0.45, above=0.05cm]{$j_{n+1}$}  node[above right] {$a_{n+1}^* = 0$};
		\path [name intersections={of=init and adapt,by=E}];
		\fill[pattern=north east lines, pattern color=red] (0, 0.3) -- (1.5, 0.75) -- (3, 0.15)
		-- (E) -- (1.5, 0.8);
		\fill[pattern=north east lines, pattern color=BOARD_GREEN] (E) -- (3.5, 0) -- (4.5, 0);
		\draw[dashed] (1.5, 1.2)  -- (1.5, -0.8) node[below] {$t_1$};
		\draw[dashed] (3, 1.2)  -- (3, -0.8) node[below] {$t_n$};
		\draw[dashed] (4.5, 1.2)  -- (4.5, -0.8) node[below] {$t_{n+1}$};
		\end{tikzpicture} 
		\subcaption{If the acceleration $a_{n+1}^*$ is limited, $a_{max, v}^*$ is shifted in such a way that the areas hatched in red and green are of equal size.}
		\label{fig:vel_limitation_b}
		
	\end{subfigure}
	\caption{Consideration of velocity limitations.}
	\label{fig:vel_limitation}
\end{figure*}
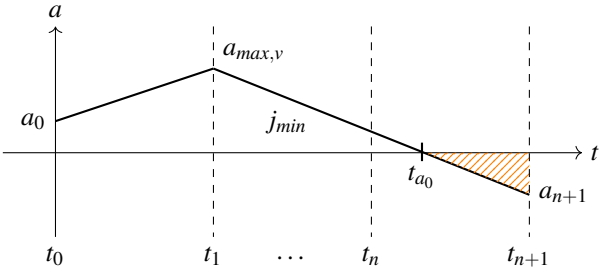
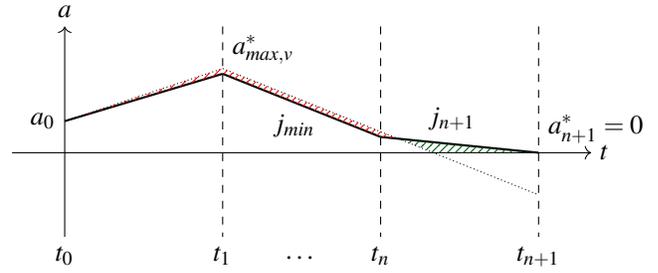

\subsubsection{Termination}
A training episode terminates if the angular deviation between $p_{t+1}$ and $p_{t+1, \; ref}$ exceeds a fixed threshold for at least one joint.
The termination serves a dual purpose: Firstly, the system learns to stay close to the reference as termination leads to a smaller sum of rewards. Secondly, violation of position limits as well as self-collision are avoided, provided that the reference trajectory maintains a certain safety distance.

\subsection{Implementation}

We use a fully-connected neural network with SELU activations \cite{klambauer2017self} and two hidden layers of size [256, 128] to map states to  actions. The training process is performed in parallel using the Ray framework \cite{DBLP:journals/corr/abs-1712-05889} and reference implementations provided by RLlib \cite{liang2017rllib}. 
Because of its stability and reliability, the on-policy algorithm PPO \cite{schulman2017proximal} is chosen for training.
The batch size is set to $2^{14}$.

\section{Consideration of Joint Limitations}
When executing online adaptations with a real robot, joint limitations have to be considered to avoid permanent damage to the robot joints. 
The basic idea of our approach is to calculate for each joint $i$ and at each decision step the acceleration range $[a_{min, i}\;, \; a_{max, i}]$ that does not lead to a violation of joint limits. As analytical expressions can be derived, the calculation can be done in real-time.
Computing the range of valid accelerations for seven joints took at most \SI{0.9}{\milli\second} with an Intel i9-9900K CPU. 
Once the valid range is known, adapting the network prediction $\underline{a_t}$ is straightforward:
\begin{align}
a_{t+1, \; i} = \begin{dcases*}
a_{min, \; i} & $\underline{a_{t, \; i}} < a_{min, \; i}$ \\
\underline{a_{t, \; i}} & $\underline{a_{t, \; i}} \in [a_{min, \; i}, \; a_{max, \; i}]$\\
a_{max, \; i} & $\underline{a_{t, \; i}} > a_{max, \; i}$
\end{dcases*}
\end{align}
$a_{max, \; i}$ is the normalized minimum value of $a_{max, j}$, $a_{max, a}$ and $a_{max, v}$, which are defined in the following. Equations for $a_{min, \; i}$ can be derived correspondingly.

\subsection{Jerk Limitation} 
Given that the jerk is constant within each control cycle, the maximum valid acceleration can be computed as follows:
\begin{align}
a_{max, j} = a_0 + j_{max} \cdot \Delta t_{NN}
\end{align}
We note that the linear interpolation of accelerations naturally limits jerk to:
\begin{align}
j_{max, \; interpolation} = \frac{a_{max} - a_{min}}{\Delta t_{NN}}
\end{align}

\subsection{Acceleration Limitation}
Restricting accelerations is trivial as the range of valid accelerations corresponds to the specified acceleration limits. 
\begin{align}
a_{max, a} = a_{max}
\end{align}
\subsection{Velocity Limitation}
To guarantee bounded velocities, it is no longer sufficient to consider the next time step only.
When working close to the velocity limit at a high acceleration, there might be no way to stay within the permitted velocity range without violating jerk limitations. Our approach prevents the robot from getting in such a situation.
Fig. \ref{fig:vel_limitation} illustrates the main idea. The maximum acceleration at the next time step $a_{max, v}$ must be followed by a deceleration with $j_{min}$. In addition,  $a_{max, v}$ has to be chosen in such a way that the maximum velocity is reached at zero acceleration. 
For ${v_0 + \frac{a_0 \cdot \Delta t_{NN} }{2} < v_{max}}$, the following formula can be derived
\begin{multline}
a_{max, v} = \\
\frac{j_{min} \cdot \Delta t_{NN}}{2} \cdot \left( 1 - \sqrt{1 + \frac{8 \cdot \left(v_0 - v_{max}\right) + 4 \cdot a_0 \cdot \Delta t_{NN}}{j_{min} \cdot \Delta t_{NN}^2}}\right), 
\end{multline}
whereas
\begin{align}
a_{max, v} = a_0 \cdot \left( 1 - \frac{1}{2} \cdot \frac{a_0 \cdot \Delta t_{NN}}{v_{max} - v_0} \right)
\end{align}
applies for $v_0 + \frac{a_0 \cdot \Delta t_{NN}}{2} \ge v_{max}$.

The approach described above can cause oscillations, as the velocity does not necessarily reach its maximum value  at a discrete decision step. In Fig. \ref{fig:vel_limitation_a}, the area hatched in orange indicates the difference between $v_{max}$ and the velocity at the next discrete decision step $v_{n+1}$. 
The problem can be mitigated by shifting $a_{max, v}$, as shown in Fig. \ref{fig:vel_limitation_b}.

\begin{figure}[h]
\vspace{0.2cm}
\centering\
\includegraphics[width=\linewidth]{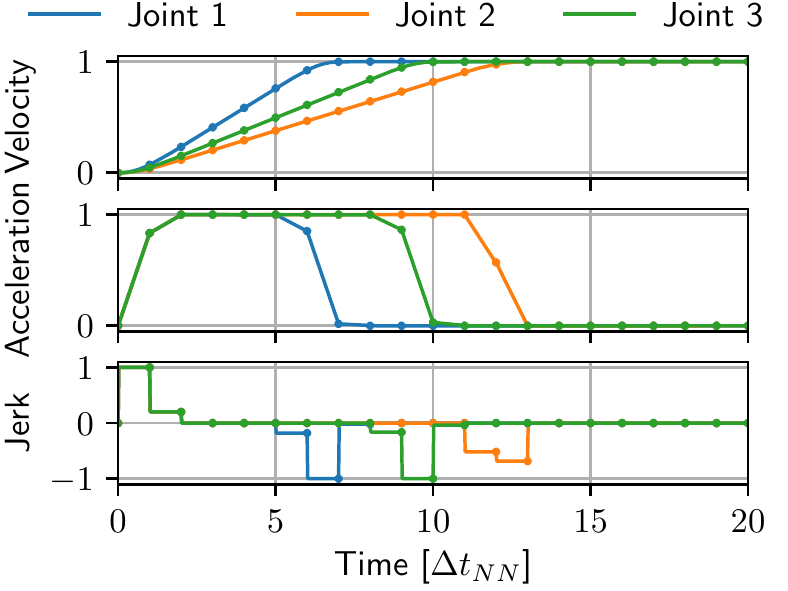}
\caption{Jerk, acceleration and velocity limitation when choosing the maximum valid acceleration at each decision step.}%
\label{fig:vel_Limitation_max_acc}
\end{figure}

\subsection{Validation}
\begin{figure}[h]
\centering\
\includegraphics[width=\linewidth]{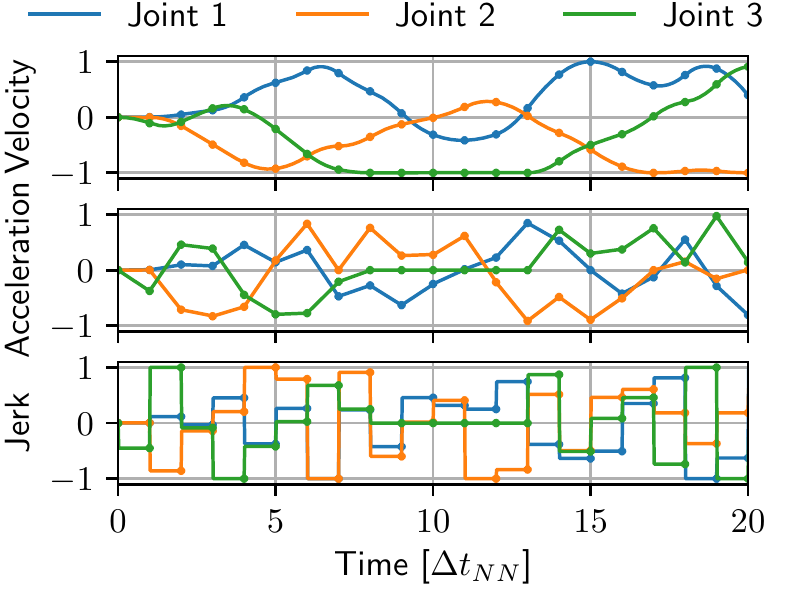}
\caption{Exemplary trajectory when choosing random accelerations within the range of valid accelerations.}%
\label{fig:vel_limitation_random_acc}
\end{figure}

We validated our approach by running tests with over \mbox{100 000} simulated trajectories without exceeding the \mbox{maximum} velocities, accelerations and jerks. 
Fig. \ref{fig:vel_Limitation_max_acc} illustrates the system behavior if the maximum acceleration is chosen at each decision step.
As expected, the acceleration is first restricted due to jerk constraints, followed by acceleration and velocity limitations.   
In Fig. \ref{fig:vel_limitation_random_acc} random accelerations are sampled from the calculated range of valid accelerations. The figure shows that smooth velocities are generated and that the joint limits are not violated.

\section{Experimental Setup}
\begin{figure*}[h]
	\begin{subfigure}[c]{0.5\textwidth}
        \includegraphics[trim=30 20 30 10, clip, width=0.95\linewidth]{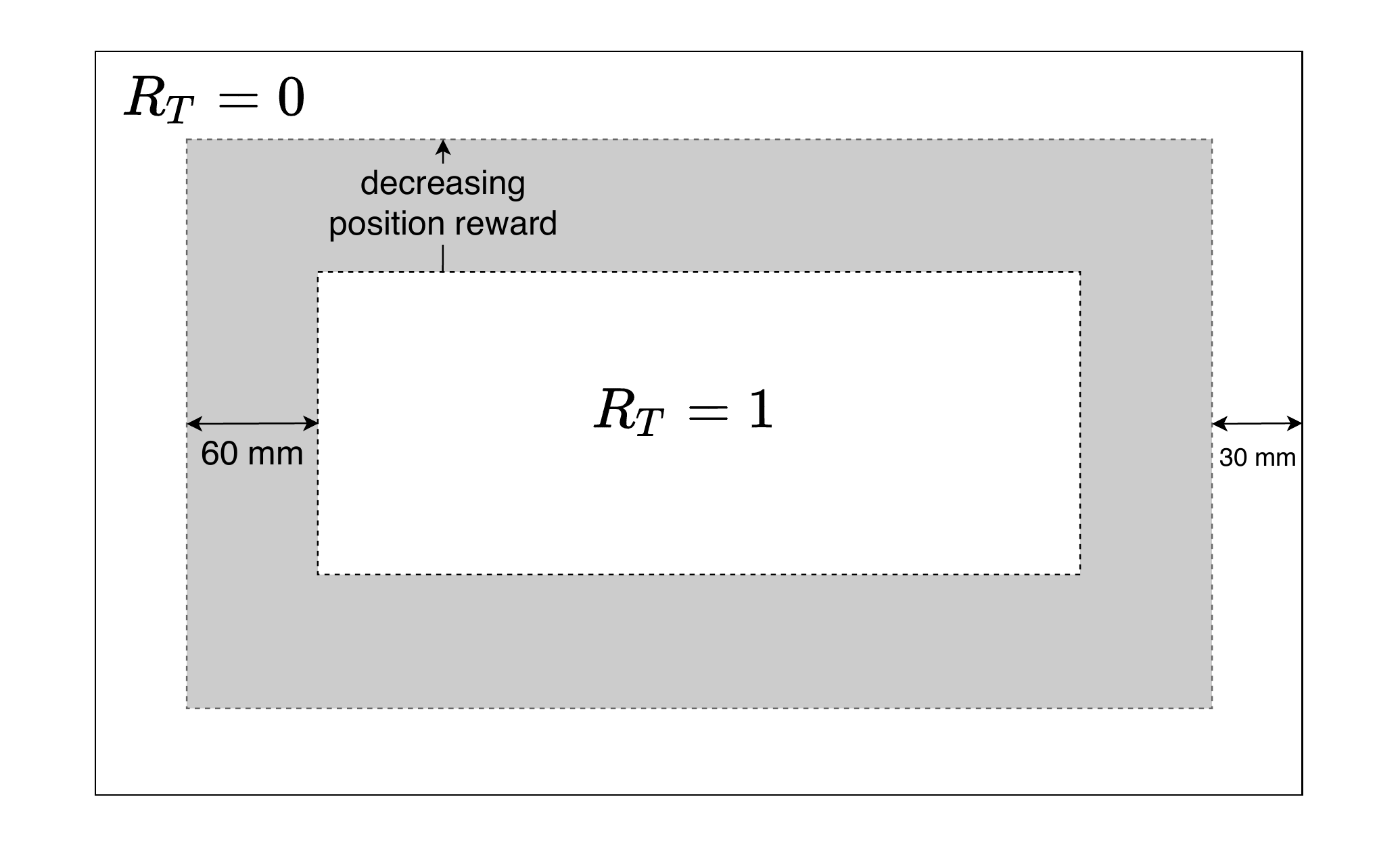} %
	    \subcaption{On plate}

	\end{subfigure}
	\begin{subfigure}[c]{0.5\textwidth}
		\includegraphics[trim=30 20 30 10, clip, width=0.95\linewidth]{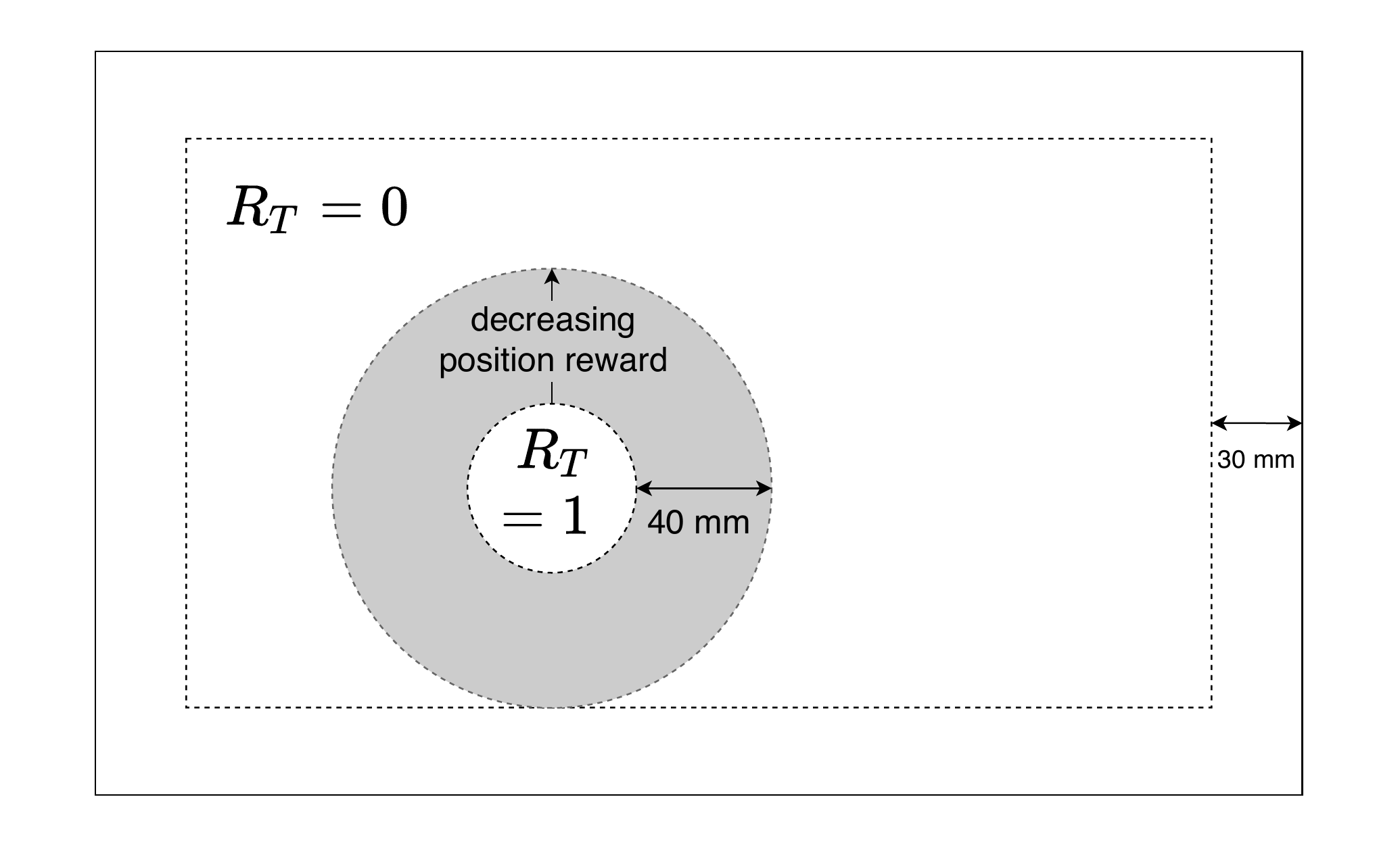} %
		\subcaption{In place}

	\end{subfigure}
	\caption{Definition of the task-specific reward $R_T$ for two versions of a ball-on-plate task.}
	\label{fig:reward_definition}
\end{figure*}

We evaluated our approach with two versions of a dynamic ball-on-plate task performed by a KUKA iiwa robot with seven degrees of freedom. While the basic task is to balance a ball on a plate during motion, the first version allows the ball to  move within a large area of the plate (``on plate''). In contrast, the second version tries to keep the ball as close as possible to its initial position (``in place''). 
Fig. \ref{fig:reward_definition} shows how the task-specific reward is defined for both cases. 
The second version is related to traditional control tasks as there is one fixed setpoint for the ball position.

\subsection{Reference Trajectories}
\begin{figure}[h]
\centering
\includegraphics[angle=-90, trim=165 0 0 0, clip, width=0.9\linewidth]{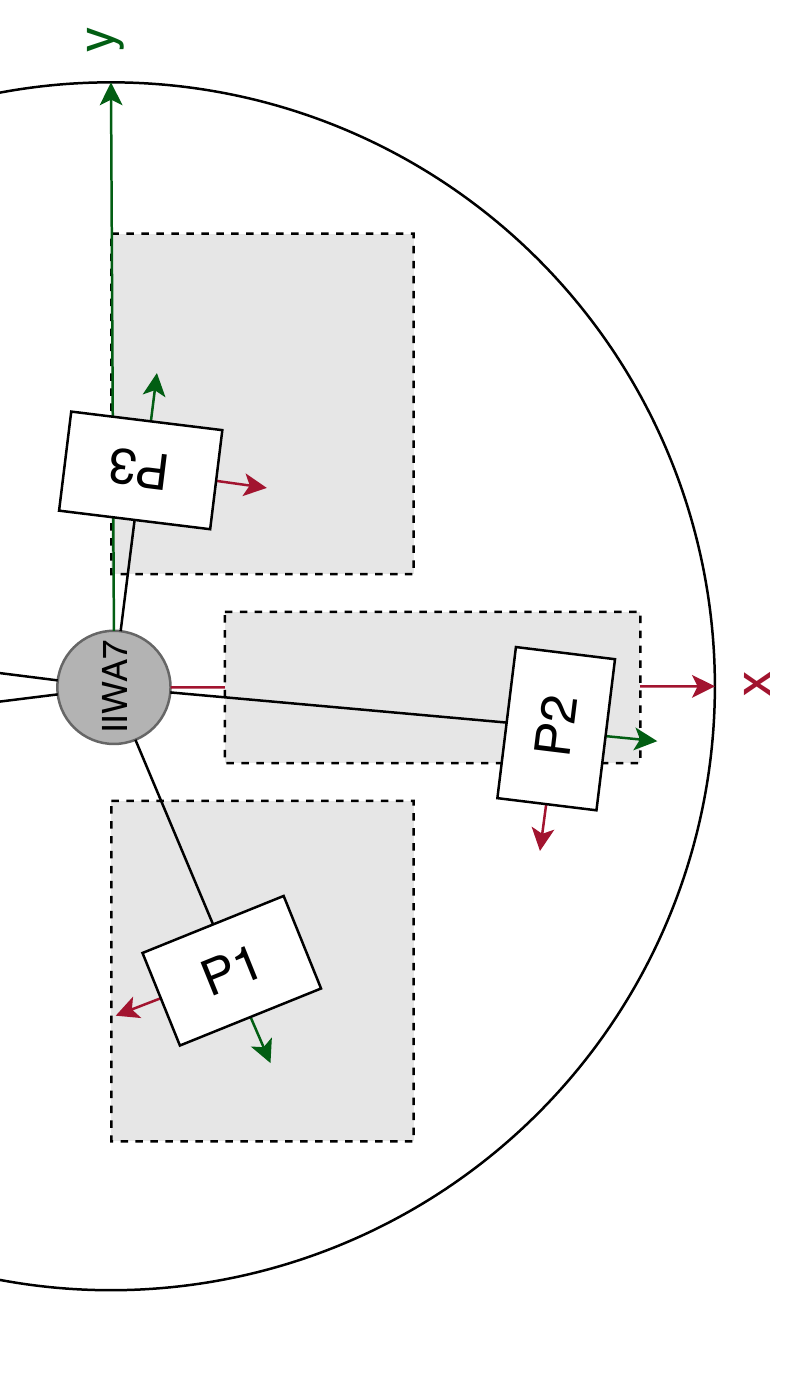}
\caption{Top view on the sampling areas to generate waypoints for an exemplary balancing task.}%
\label{fig:Initial_Sampling}
\end{figure}

The training dataset consists of 150 000 reference trajectories at different heights with sampling areas like those shown in Fig. \ref{fig:Initial_Sampling}. For reasons of symmetry, each trajectory can be mirrored along two planes, leading to a total of 600 000 trajectories. 

\subsection{Sensory Feedback}
Feedback on the task execution is given by adding the current and the last ball position to the state.  
For the ``in place'' task, we additionally include the two-dimensional distance to the initial ball position, which serves as a measure of the control error. 

\begin{figure}[h]
\centering
\includegraphics[trim=50 350 400 450, clip, width=1.0\linewidth]{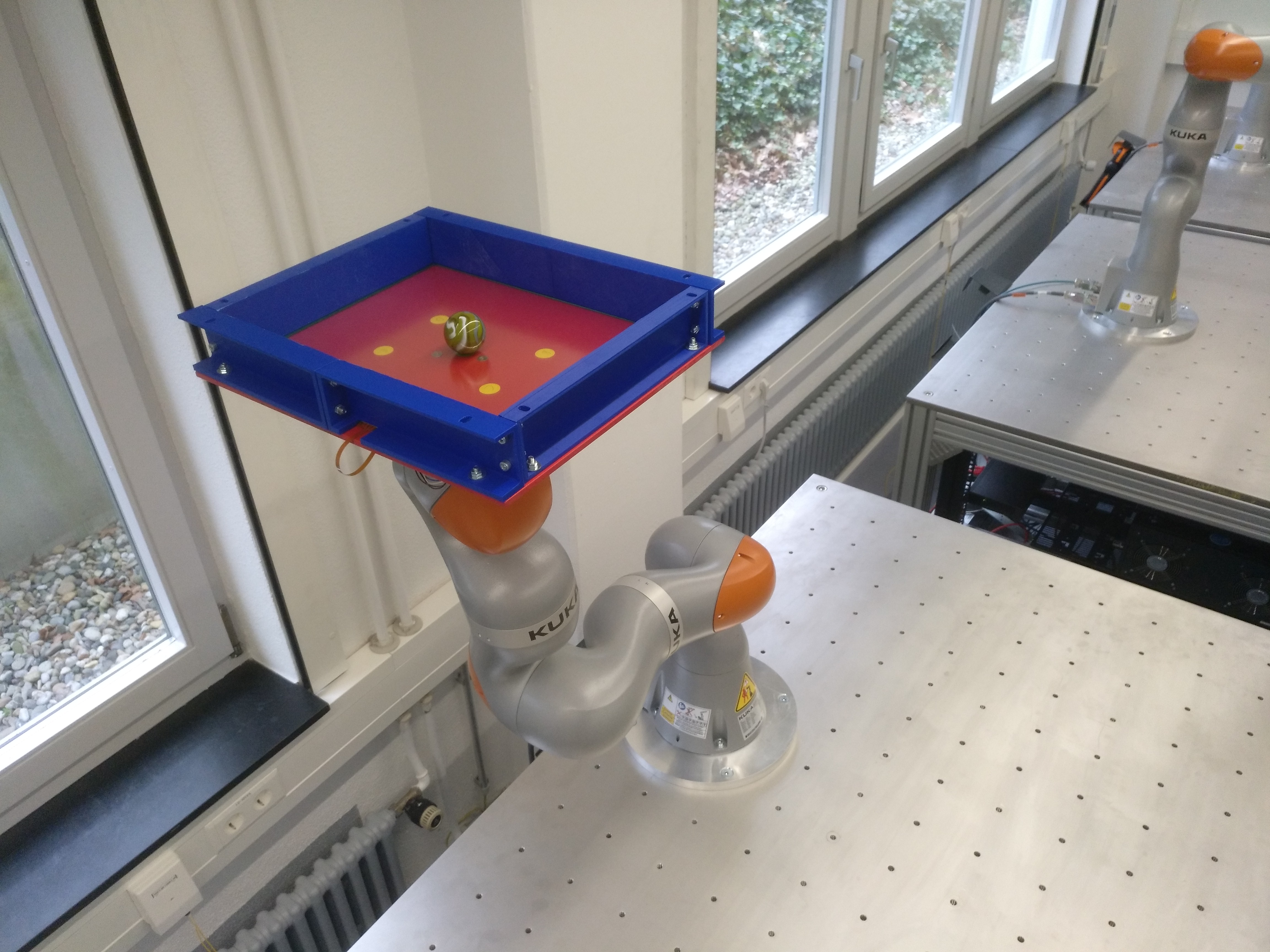}
\caption{Real-world setup for sim-to-real transfer.}%
\label{fig:Sim2Real_Setup}
\end{figure}

\subsection{Physics Simulation}
The physics engine PyBullet \cite{coumans2016pybullet} is used to generate training data in simulation. With the aim to learn a robust policy, we randomize the ball characteristics (mass, friction, radius) and model the measuring error of the ball position by adding noise to the corresponding signal.

\begin{table*}[h]

\centering
\begin{tabular}{cl c c c c c c}
\hline
& Setting & Success rate & Trajectory fraction & Error distance  & Acceleration  & Jerk \\
\hline
& Reference trajectories (no adaptations) & \SI{4.2}{\percent} & \SI{40.4}{\percent}  & - & \SI{1.7}{\percent} & \SI{0.5}{\percent}\\
& True\AE dapt: test set & \SI{89.6}{\percent} & \SI{97.6}{\percent}  & - & \SI{7.0}{\percent} & \SI{3.7}{\percent}\\
& True\AE dapt: training set & \SI{90.7}{\percent} & \SI{97.9}{\percent}  & - & \SI{7.0}{\percent} & \SI{3.7}{\percent}\\
& Open loop: evaluation only &   \SI{8.9}{\percent} & \SI{53.4}{\percent}  & - & \SI{7.0}{\percent} & \SI{3.6}{\percent}\\
& Open loop: training and evaluation &   \SI{45.6}{\percent} & \SI{84.2}{\percent}  & - & \SI{7.1}{\percent} & \SI{4.1}{\percent}\\
& State: no current position &   \SI{0.1}{\percent} & \SI{30.8}{\percent}  & - & \SI{7.4}{\percent} & \SI{4.5}{\percent}\\
& State: no current velocity &   \SI{13.7}{\percent} & \SI{68.5}{\percent}  & - & \SI{5.4}{\percent} & \SI{3.2}{\percent}\\
& State: no current acceleration &   \SI{91.4}{\percent} & \SI{97.9}{\percent}  & - & \SI{20.8}{\percent} & \SI{16.0}{\percent}\\
& State: no following positions  &   \SI{0.4}{\percent} & \SI{31.3}{\percent}  & - & \SI{8.8}{\percent} & \SI{5.4}{\percent}\\
& State: ten following positions &   \SI{92.9}{\percent} & \SI{98.3}{\percent}  & - & \SI{20.2}{\percent} & \SI{15.5}{\percent}\\
& Punishment: no jerk penalty &   \SI{87.5}{\percent} & \SI{96.7}{\percent}  & - & \SI{23.0}{\percent} & \SI{18.1}{\percent}\\
\rot{\rlap{On plate}}
& Punishment: no acceleration penalty &   \SI{50.6}{\percent} & \SI{79.3}{\percent}  & - & \SI{62.4}{\percent} & \SI{50.4}{\percent}\\

\hline
& Reference trajectories (no adaptations) & \SI{0.3}{\percent} & \SI{22.1}{\percent}  & - & \SI{2.0}{\percent} & \SI{0.3}{\percent}\\
& True\AE dapt: test set & \SI{98.6}{\percent} & \SI{99.7}{\percent}  & \SI{1.2}{\centi\meter} & \SI{6.4}{\percent} & \SI{3.4}{\percent}\\
& True\AE dapt: real robot & \SI{82.0}{\percent} & \SI{96.1}{\percent}  & \SI{1.7}{\centi\meter} & \SI{6.9}{\percent} & \SI{3.6}{\percent}\\
& Open loop: evaluation only & \SI{34.3}{\percent} & \SI{73.0}{\percent}  & \SI{2.4}{\centi\meter} & \SI{6.6}{\percent} & \SI{3.3}{\percent}\\
\rot{\rlap{In place}}
& Open loop: training and evaluation & \SI{61.7}{\percent} & \SI{91.7}{\percent}  & \SI{1.6}{\centi\meter} & \SI{6.7}{\percent} & \SI{4.1}{\percent}\\
\hline
\end{tabular}
\caption{Average values of the success rate, the successful trajectory fraction and the distance to the initial ball position for different configurations. The mean of the normalized absolute accelerations and jerks is averaged over all joints.}
\label{table:evalution}
\end{table*}

\subsection{Real Setup}
A picture of the real setup is shown in Fig. \ref{fig:Sim2Real_Setup}.
For real-world experiments, the current ball position is detected by a resistive touch panel with a size of \SI{34}{\centi\meter} $\times$ \SI{27}{\centi\meter}. The robot is controlled via position commands at a rate of \SI{200}{\hertz}.

\section{Evaluation}
We define two metrics to measure the performance of the task execution, namely the success rate and the successfully executed trajectory fraction. For the ``on plate'' task, a trajectory is considered as successful if the ball does not touch the border of the plate, while the ``in place'' task allows a deviation of at most \SI{6}{\centi\meter} from the initial ball position.   
The performance of the ``on plate'' task is evaluated after 32 million training steps, whereas 220 million training steps were conducted for the ``in place'' task. To generate the performance metrics in simulation, \mbox{10 000} trajectories from the test set were executed. Real world performance was evaluated with 50 trajectories and five different initial ball positions as indicated by the yellow spots in Fig. \ref{fig:Sim2Real_Setup}. For the ``on plate'' task, a trajectory fraction of \SI{97.6}{\percent} and a success rate of \SI{89.6}{\percent} was achieved. The ``in place'' task accomplished a trajectory fraction of \SI{99.7}{\percent} and a success rate of \SI{98.6}{\percent}. Transferring the policy to a real robot led to a a trajectory fraction of \SI{96.1}{\percent} and a success rate of \SI{82.0}{\percent}.

\subsection{Ablation Studies}
Ablation studies were performed to analyze the influence of individual system components. The results are listed in TABLE \ref{table:evalution}.
As expected, the network was not able to learn the task when omitting the current position or the next position of the reference trajectory from the state. Poor performance was achieved when omitting the current velocity.  
Our experiments show that the acceleration penalty is crucial for successful task execution. Without the penalty, jerky movements are produced, making it potentially harder to control the ball. The jerk penalty further improves the smoothness of the generated trajectories. Adding more than one future reference position to the state had a marginal impact on the performance. However, having access to more points might be crucial for tasks with a longer planning horizon.

\subsection{Importance of Sensory Feedback}
\begin{figure}[h]
\vspace{0.2cm}
\centering\begin{tabular}{c}
\includegraphics[trim=600 670 510 80,clip,width=\linewidth]{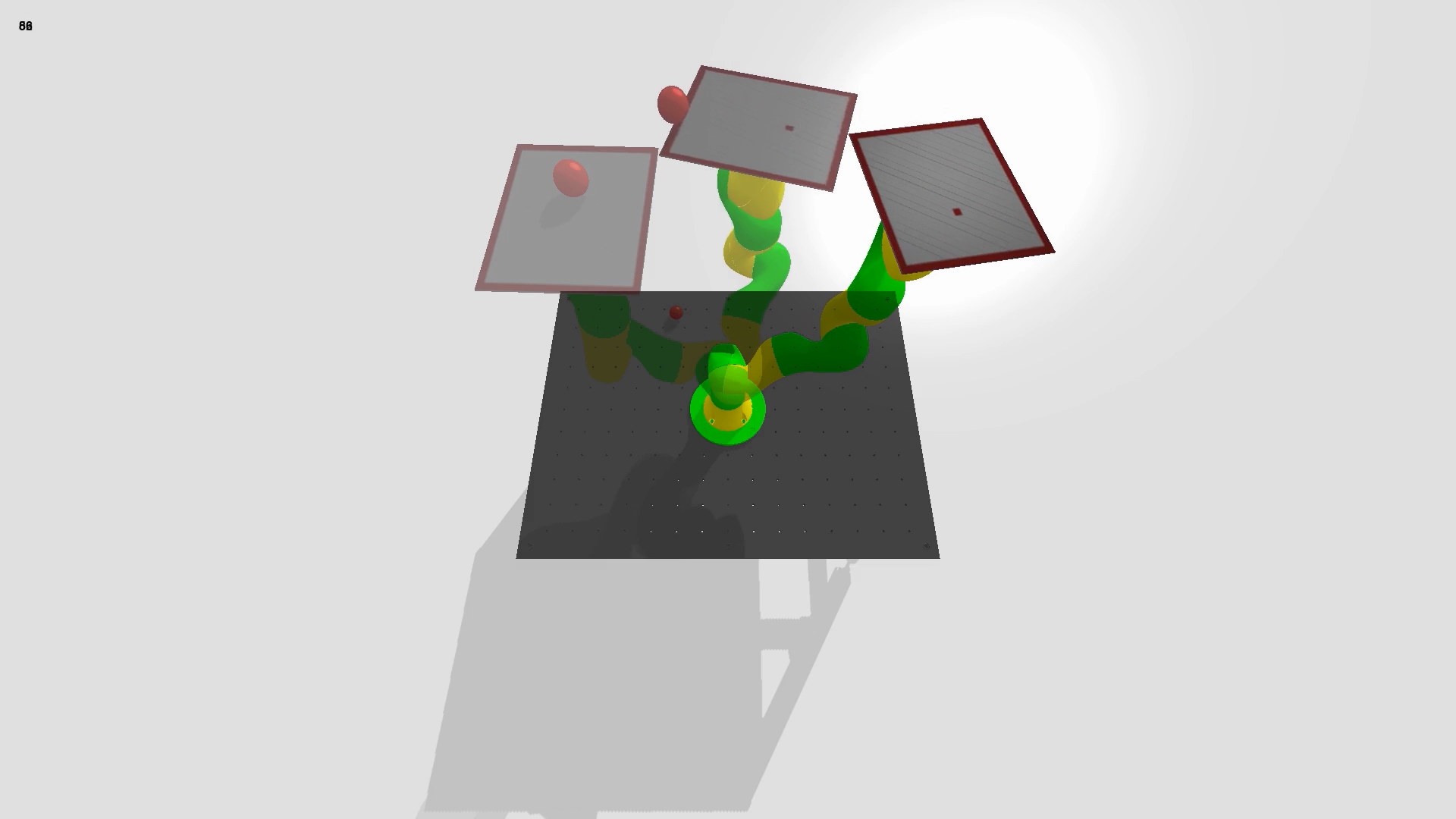}\\
\includegraphics[trim=600 670 510 80, clip, width=\linewidth]{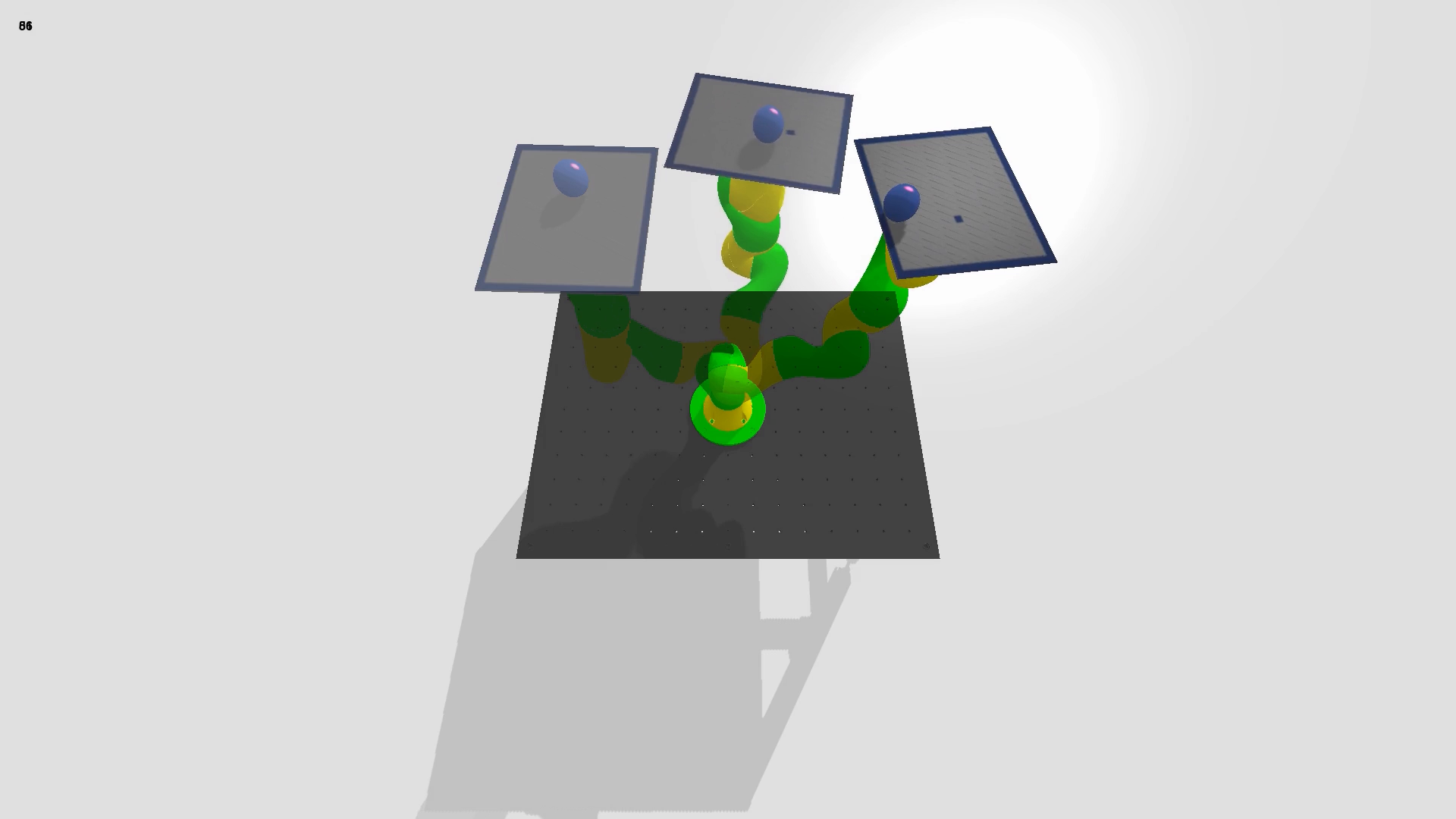}\\
\end{tabular}
\caption{Exemplary trajectory execution.
Top: The reference trajectory fails to keep the ball on the plate.  Bottom: When adapting the trajectory without updating the sensor signals, the ball stays on the plate but not at the desired spot.}%
\label{fig:Intial_Open_Loop_Dynamic}
\end{figure}

To assess the importance of closed loop feedback, we analyzed the performance of a network trained with sensory feedback without updating the sensor signals during evaluation (open loop). For the ``in place'' task the success rate dropped from \SI{98.6}{\percent} to \SI{34.3}{\percent}, showing that the feedback is essential for the network. 
An exemplary rollout is illustrated in Fig. \ref{fig:Intial_Open_Loop_Dynamic}.  Training a policy from scratch without sensory feedback led to a success rate of \SI{61.7}{\percent}. We conclude that the network has, to a certain extent, learned to anticipate future movements of the ball.

\begin{figure}[h]
\centering
\includegraphics[width=1.0\linewidth]{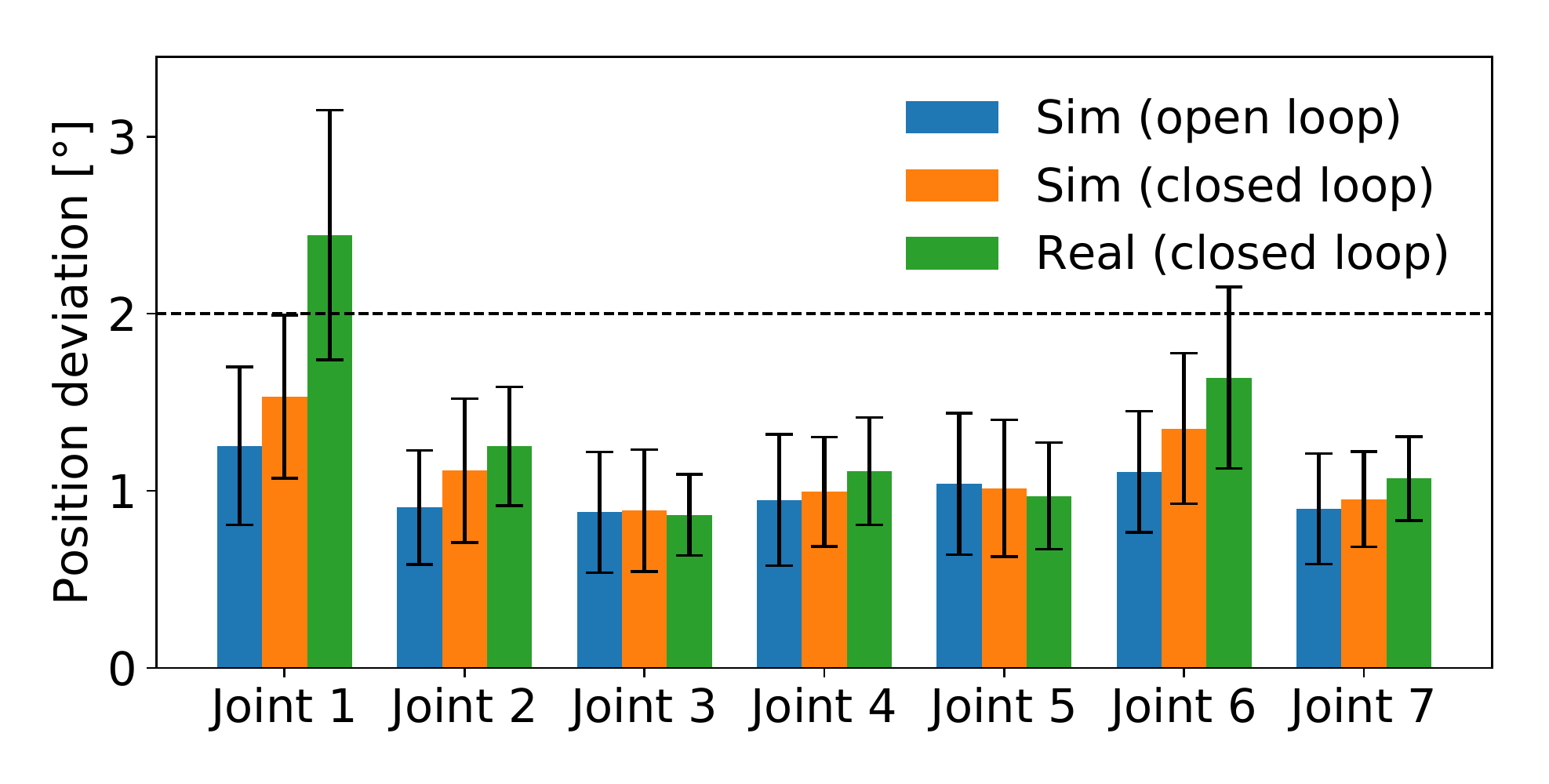}
\caption{Mean position deviation for the ``in place'' task. The lower threshold of the deviation penalty $\Delta p_{l}$ was set to \SI{2}{\degree}.}%
\label{fig:Sim2Real_Pos_Deviation}
\end{figure}

\subsection{Deviation to the Reference Trajectory}
Fig. \ref{fig:Sim2Real_Pos_Deviation} shows the mean position deviation from the reference trajectory for the ``in place'' task. On average, all joints stay close to their reference. During real-world execution, stronger adaptations are predicted, especially for joint 1 and joint 6. This appears reasonable as the movement of the ball is harder to predict if the target domain differs from the training domain.

\subsection{Sim-to-Real Transfer}
\begin{figure}[h]
\centering
\includegraphics[width=1.0\linewidth]{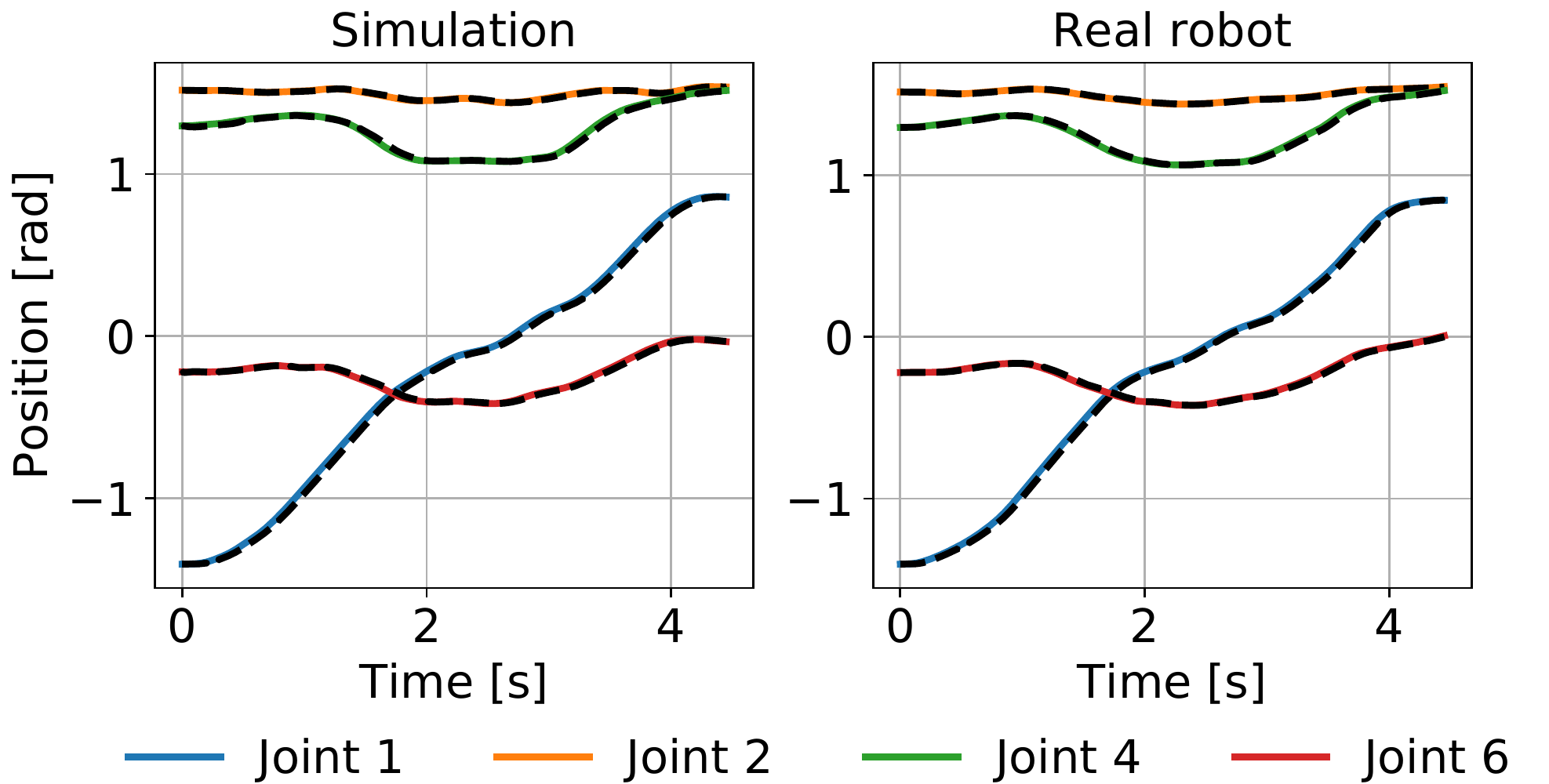}
\caption{Tracking performance of the trajectory controller in simulation and in the real world. Setpoints are shown as solid lines. Actual values are represented by dashed black lines.}%
\label{fig:sim2real_Actual_Values}
\end{figure}

Assuming that the actual joint positions closely follow their setpoints, we use setpoints instead of actual values for the robot state. Fig. \ref{fig:sim2real_Actual_Values} visualizes the tracking accuracy of the trajectory controller in simulation and in the real world. During fast movements a small delay can be noticed. However, as the delay appears in both simulated and real data, the policy can learn to cope with it.

\begin{figure}[h]
\centering
\includegraphics[width=\linewidth]{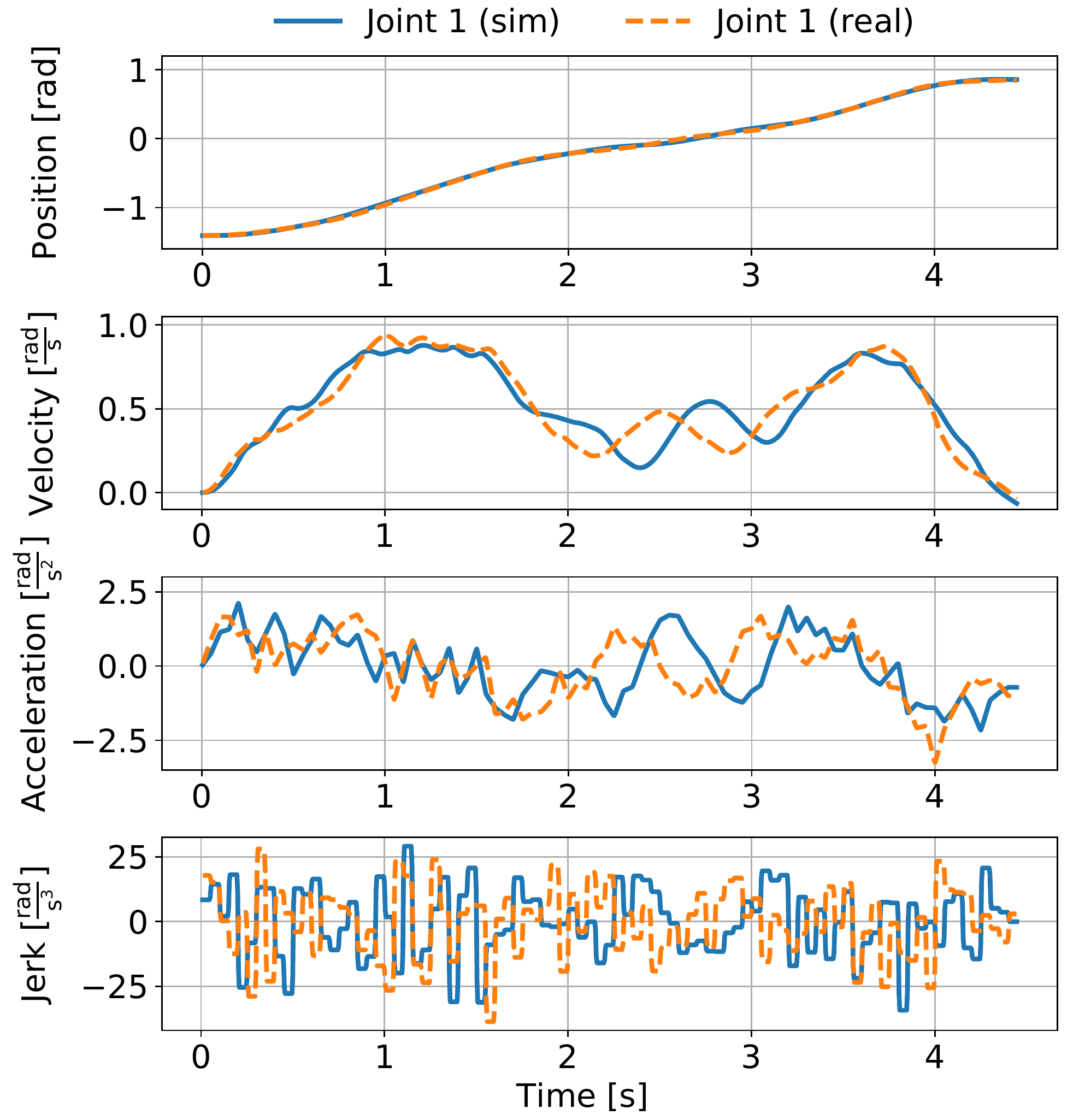}
\caption{Comparison of simulated and real setpoints for an exemplary trajectory execution of the ``in place'' task.}%
\label{fig:sim2real_setpoints}
\end{figure}

Fig. \ref{fig:sim2real_setpoints} shows a successful rollout of the ``in place'' task for both simulation and real execution.  

As shown in the accompanying video, the sim-to-real transfer could be successfully conducted with various balls, differing in mass, size and material.

\section{CONCLUSIONS}
We presented a real-time capable approach for learning online adaptations based on sensory feedback and a method to ensure that the jerks, accelerations and velocities of the adapted trajectories are bounded. The effectiveness of our approach was demonstrated by learning to balance a ball on a plate while moving. The policy was trained in simulation and successfully transferred to a real robot. 
The evaluation showed that the adapted trajectories stay close to their reference and that sensory feedback is crucial for successful task execution. In future work, we intend to analyze the performance of our approach for tasks that require further deviation from the reference trajectory. In addition, we aim to develop a more sophisticated method for avoiding position limits and self-collision during motion.

\section*{ACKNOWLEDGMENT}

This research was supported by the German Federal Ministry of Education and Research (BMBF) and the Indo-German Science \& Technology Centre (IGSTC) as part of the project TransLearn (01DQ19007A). We would like to thank Tamim Asfour for his valuable input and helpful advice. All real-world experiments were performed at the KUKA Robot Learning Lab at KIT \cite{RLL}. Special thanks to Wolfgang Wiedmeyer for his tremendous effort in building up the lab.

\bibliographystyle{IEEEtran}
\bibliography{root}

\begin{thebibliography}{10}
\providecommand{\url}[1]{#1}
\csname url@rmstyle\endcsname
\providecommand{\newblock}{\relax}
\providecommand{\bibinfo}[2]{#2}
\providecommand\BIBentrySTDinterwordspacing{\spaceskip=0pt\relax}
\providecommand\BIBentryALTinterwordstretchfactor{4}
\providecommand\BIBentryALTinterwordspacing{\spaceskip=\fontdimen2\font plus
\BIBentryALTinterwordstretchfactor\fontdimen3\font minus
  \fontdimen4\font\relax}
\providecommand\BIBforeignlanguage[2]{{%
\expandafter\ifx\csname l@#1\endcsname\relax
\typeout{** WARNING: IEEEtran.bst: No hyphenation pattern has been}%
\typeout{** loaded for the language `#1'. Using the pattern for}%
\typeout{** the default language instead.}%
\else
\language=\csname l@#1\endcsname
\fi
#2}}

\bibitem{geraerts2004comparative}
R.~Geraerts and M.~H. Overmars, ``A comparative study of probabilistic roadmap
  planners,'' in \emph{Algorithmic Foundations of Robotics V}.\hskip 1em plus
  0.5em minus 0.4em\relax Springer, 2004, pp. 43--57.

\bibitem{kunz2012time}
T.~Kunz and M.~Stilman, ``Time-optimal trajectory generation for path following
  with bounded acceleration and velocity,'' \emph{Robotics: Science and Systems
  VIII}, pp. 1--8, 2012.

\bibitem{Reflexxes}
T.~{Kröger}, ``Opening the door to new sensor-based robot applications—the
  reflexxes motion libraries,'' in \emph{2011 IEEE International Conference on
  Robotics and Automation}, May 2011, pp. 1--4.

\bibitem{toppra}
H.~Pham and Q.~C. Pham, ``A new approach to time-optimal path parameterization
  based on reachability analysis,'' \emph{IEEE Transactions on Robotics},
  vol.~34, pp. 645 -- 659, 06 2018.

\bibitem{DBLP:journals/corr/abs-1812-11103}
\BIBentryALTinterwordspacing
T.~Haarnoja, A.~Zhou, S.~Ha, J.~Tan, G.~Tucker, and S.~Levine, ``Learning to
  walk via deep reinforcement learning,'' \emph{CoRR}, vol. abs/1812.11103,
  2018. [Online]. Available: \url{http://arxiv.org/abs/1812.11103}
\BIBentrySTDinterwordspacing

\bibitem{tan2018sim}
J.~Tan, T.~Zhang, E.~Coumans, A.~Iscen, Y.~Bai, D.~Hafner, S.~Bohez, and
  V.~Vanhoucke, ``Sim-to-real: Learning agile locomotion for quadruped
  robots,'' \emph{arXiv preprint arXiv:1804.10332}, 2018.

\bibitem{berscheid_icra19}
L.~{Berscheid}, T.~{Rühr}, and T.~{Kröger}, ``Improving data efficiency of
  self-supervised learning for robotic grasping,'' in \emph{2019 International
  Conference on Robotics and Automation (ICRA)}, May 2019, pp. 2125--2131.

\bibitem{kalashnikov2018qt}
D.~Kalashnikov, A.~Irpan, P.~Pastor, J.~Ibarz, A.~Herzog, E.~Jang, D.~Quillen,
  E.~Holly, M.~Kalakrishnan, V.~Vanhoucke, \emph{et~al.}, ``Qt-opt: Scalable
  deep reinforcement learning for vision-based robotic manipulation,''
  \emph{arXiv preprint arXiv:1806.10293}, 2018.

\bibitem{nagabandi2019deep}
A.~Nagabandi, K.~Konoglie, S.~Levine, and V.~Kumar, ``Deep dynamics models for
  learning dexterous manipulation,'' \emph{arXiv preprint arXiv:1909.11652},
  2019.

\bibitem{DBLP:journals_OpenAI_In_Hand}
\BIBentryALTinterwordspacing
OpenAI, M.~Andrychowicz, B.~Baker, M.~Chociej, R.~J{\'{o}}zefowicz, B.~McGrew,
  J.~W. Pachocki, J.~Pachocki, A.~Petron, M.~Plappert, G.~Powell, A.~Ray,
  J.~Schneider, S.~Sidor, J.~Tobin, P.~Welinder, L.~Weng, and W.~Zaremba,
  ``Learning dexterous in-hand manipulation,'' \emph{CoRR}, vol.
  abs/1808.00177, 2018. [Online]. Available:
  \url{http://arxiv.org/abs/1808.00177}
\BIBentrySTDinterwordspacing

\bibitem{TrajectoryOptimizationReferencePath}
K.~{Ota}, D.~K. {Jha}, T.~{Oiki}, M.~{Miura}, T.~{Nammoto}, D.~{Nikovski}, and
  T.~{Mariyama}, ``Trajectory optimization for unknown constrained systems
  using reinforcement learning,'' in \emph{2019 IEEE/RSJ International
  Conference on Intelligent Robots and Systems (IROS)}, Nov 2019, pp.
  3487--3494.

\bibitem{levine2016end}
S.~Levine, C.~Finn, T.~Darrell, and P.~Abbeel, ``End-to-end training of deep
  visuomotor policies,'' \emph{The Journal of Machine Learning Research},
  vol.~17, no.~1, pp. 1334--1373, 2016.

\bibitem{tobin2017domain}
J.~Tobin, R.~Fong, A.~Ray, J.~Schneider, W.~Zaremba, and P.~Abbeel, ``Domain
  randomization for transferring deep neural networks from simulation to the
  real world,'' in \emph{2017 IEEE/RSJ international conference on intelligent
  robots and systems (IROS)}.\hskip 1em plus 0.5em minus 0.4em\relax IEEE,
  2017, pp. 23--30.

\bibitem{peng2018sim}
X.~B. Peng, M.~Andrychowicz, W.~Zaremba, and P.~Abbeel, ``Sim-to-real transfer
  of robotic control with dynamics randomization,'' in \emph{2018 IEEE
  international conference on robotics and automation (ICRA)}.\hskip 1em plus
  0.5em minus 0.4em\relax IEEE, 2018, pp. 1--8.

\bibitem{graspGAN}
K.~Bousmalis, A.~Irpan, P.~Wohlhart, Y.~Bai, M.~Kelcey, M.~Kalakrishnan,
  L.~Downs, J.~Ibarz, P.~Pastor, K.~Konolige, \emph{et~al.}, ``Using simulation
  and domain adaptation to improve efficiency of deep robotic grasping,'' in
  \emph{2018 IEEE International Conference on Robotics and Automation
  (ICRA)}.\hskip 1em plus 0.5em minus 0.4em\relax IEEE, 2018, pp. 4243--4250.

\bibitem{TrueRMA}
J.~C. Kiemel, P.~Meißner, and T.~Kröger, ``{TrueRMA:} {Learning} fast and
  smooth robot trajectories with recursive midpoint adaptations in cartesian
  space,'' in \emph{{International} {Conference} on Robotics and {Automation}
  ({ICRA})}.\hskip 1em plus 0.5em minus 0.4em\relax IEEE, 2020.

\bibitem{kalakrishnan2011stomp}
M.~Kalakrishnan, S.~Chitta, E.~Theodorou, P.~Pastor, and S.~Schaal, ``Stomp:
  Stochastic trajectory optimization for motion planning,'' in \emph{2011 IEEE
  international conference on robotics and automation}.\hskip 1em plus 0.5em
  minus 0.4em\relax IEEE, 2011, pp. 4569--4574.

\bibitem{klambauer2017self}
G.~Klambauer, T.~Unterthiner, A.~Mayr, and S.~Hochreiter, ``Self-normalizing
  neural networks,'' in \emph{Advances in neural information processing
  systems}, 2017, pp. 971--980.

\bibitem{DBLP:journals/corr/abs-1712-05889}
\BIBentryALTinterwordspacing
P.~Moritz, R.~Nishihara, S.~Wang, A.~Tumanov, R.~Liaw, E.~Liang, W.~Paul, M.~I.
  Jordan, and I.~Stoica, ``Ray: {A} distributed framework for emerging {AI}
  applications,'' \emph{CoRR}, vol. abs/1712.05889, 2017. [Online]. Available:
  \url{http://arxiv.org/abs/1712.05889}
\BIBentrySTDinterwordspacing

\bibitem{liang2017rllib}
E.~Liang, R.~Liaw, P.~Moritz, R.~Nishihara, R.~Fox, K.~Goldberg, J.~E.
  Gonzalez, M.~I. Jordan, and I.~Stoica, ``Rllib: Abstractions for distributed
  reinforcement learning,'' \emph{arXiv preprint arXiv:1712.09381}, 2017.

\bibitem{schulman2017proximal}
J.~Schulman, F.~Wolski, P.~Dhariwal, A.~Radford, and O.~Klimov, ``Proximal
  policy optimization algorithms,'' \emph{arXiv preprint arXiv:1707.06347},
  2017.

\bibitem{coumans2016pybullet}
E.~Coumans and Y.~Bai, ``Pybullet, a python module for physics simulation for
  games, robotics and machine learning,'' \emph{GitHub repository}, 2016.

\bibitem{RLL}
W.~{Wiedmeyer}, M.~{Mende}, D.~{Hartmann}, R.~{Bischoff}, C.~{Ledermann}, and
  T.~{Kroger}, ``Robotics education and research at scale: A remotely
  accessible robotics development platform,'' in \emph{2019 International
  Conference on Robotics and Automation (ICRA)}, May 2019, pp. 3679--3685.

\end{thebibliography}

\end{document}